\documentclass[runningheads]{llncs}
\usepackage[T1]{fontenc}
\usepackage{newtxtext}
\usepackage[varvw]{newtxmath}
\usepackage{booktabs}
\usepackage{amsmath}
\usepackage{xcolor}
\usepackage{tikz}
\usetikzlibrary{positioning,arrows.meta,fit}
\usepackage{pgfplots}
\pgfplotsset{compat=1.17}
\makeatletter
\@ifundefined{credits}{\newenvironment{credits}{\par\medskip}{\par}}{}
\providecommand{\ackname}{Acknowledgements}

\makeatother

\usepackage[colorlinks=true,linkcolor=blue,citecolor=blue,urlcolor=blue,bookmarks=true]{hyperref}

\begin{document}

\title{Forecast Value Add: Reward-Based Action Selection over Probabilistic Demand Forecasts}
\titlerunning{Forecast Value Add via Reward-Based Action Selection}

\author{Ankur Verma}
\authorrunning{A. Verma et al.}
\institute{TrueGradient AI, \email{ankur@truegradient.ai}}

\maketitle
\begin{abstract}
Strong machine-learning ensembles capture much of the trend, seasonality, and known events in demand, yet they can still understate peaks, miss moving events, or mistake temporary changes for shifts in demand level. We address this gap by framing forecast refinement as a reward-driven decision problem. For each forecast period, we let an agent choose among values already produced by the ensemble: its base forecast, distribution-free bounds, quantiles, and a bias-corrected estimate. We evaluate these choices against context-matched historical periods, accounting for calendar position, recurring and moving events, trend, price, and promotions. We then apply a structural seasonal guard to constrain adjustments and record the model, agent, and planner contributions to every published value. We evaluate our approach across five retail, consumer-goods, and direct-to-consumer deployments over nine months. Our refined forecasts reduce mean WMAPE by 6.3\% compared with the stacked machine-learning forecast for every customer, with gains reaching 16.7\%, and achieve the lowest error in 35 of 45 customer-months. Compared with statistical forecasts, our approach reduces mean error by 18.7\%. These results demonstrate that context-aware, auditable action selection can add forecast value beyond a strong ensemble.

\keywords{Probabilistic Machine Learning \and Reinforcement Learning \and Planner Agents \and Forecast Value Addition}
\end{abstract}

\section{Introduction}

Demand planning is a decision problem layered on a prediction problem: a statistical engine answers ``what is the most likely quantity?'', a planner ``what quantity should we commit against?''.

\paragraph{What the ensemble already learns.}
A well-engineered machine-learning stack is not blind to structure. Fourier terms, calendar ordinals and seasonal lags give it the seasonal profile; a trend index and rolling means give it the level and its drift; holiday dummies, promotion flags and price drivers give it recurring events and commercial effects. The question is whether a separate decision layer can \emph{add value on top of it}.

\paragraph{Where value remains.}
Three kinds of residual survive even a strong ensemble. (i)~\emph{Seasonal amplitude}: an average-loss objective learns the shape of the season but slightly compresses its peaks and troughs, most in the highest-margin periods. (ii)~\emph{Rare and moving events}: a one-off promotion or a Black Friday that moved into another reporting month has few examples and a flag collinear with calendar features, so its lift is only partly attributed. (iii)~\emph{Level shifts}: a soft period, stock-out or data gap is partly absorbed into trend. These residuals recur at the same calendar positions, which makes them correctable by a policy that looks at the same positions in the past.

\paragraph{Forecast Value Add as the objective.}
Forecasting research scores models on accuracy; planning research scores \emph{interventions} against a baseline using Forecast Value Add (FVA)~\cite{ref_gilliland,ref_morlidge}. FVA asks whether a step in the process improved on what preceded it, which is exactly what an agent that modifies an existing forecast is responsible for. We pose the refinement as decision making over a \emph{pre-sanctioned, discrete} action set drawn from the ensemble's own probabilistic outputs, so each figure is traceable and the policy is inspectable as ``which action, and why''.

\paragraph{Contributions.}
(1)~A distribution-free action set from an AutoML stack and Chebyshev--Cantelli bounds (Sect.~\ref{sec:actions}); (2)~an attribution whose layers sum exactly to the model output (Sect.~\ref{sec:factor}); (3)~a per-step reward from context-matched past periods, guarded by a structural seasonal model (Sect.~\ref{sec:rl}); (4)~a worked episode for one peak row (Sect.~\ref{sec:example}); (5)~an evaluation on five customers over nine months, including the months the agent does \emph{not} win (Sect.~\ref{sec:results}).

\section{Related Work}

\paragraph{Forecasting.}
The M5 competition showed that boosted trees over engineered calendar and price features remain highly competitive and that combinations dominate single learners~\cite{ref_m5acc}; DeepAR~\cite{ref_deepar}, N-BEATS~\cite{ref_nbeats} and the Temporal Fusion Transformer~\cite{ref_tft} exploit cross-series structure. Probabilistic forecasting frames the task as distribution estimation~\cite{ref_m5unc,ref_gneiting}, with quantiles estimated directly~\cite{ref_koenker,ref_mqrnn}. We instead derive distribution-free bounds from residual dispersion, trading sharpness for validity.

\paragraph{Adjustment, explanation and decisions.}
Reconciliation aligns forecasts across a hierarchy~\cite{ref_hts}, while the judgemental-adjustment literature finds that planner overrides often destroy accuracy and increase bias~\cite{ref_morlidge}, motivating our FVA framing and governance constraints. Unlike SHAP~\cite{ref_shap}, our attribution is a deterministic counterfactual that reconciles exactly to the published number. Contextual bandits~\cite{ref_bandit} and reinforcement learning~\cite{ref_sutton} supply the vocabulary for choosing among finite actions under a reward.

\section{Problem Formulation}

Let $i$ index a product--location series and $t$ a period with seasonal cycle length $m\in\{12,52,365\}$; $a_{i,t}$ is realised demand and $\hat y_{i,t}$ the base ensemble forecast. Accuracy is measured by weighted absolute percentage error over an evaluation set $\mathcal E$,
\begin{equation}
\mathrm{WMAPE}(\mathcal E)=\frac{\sum_{(i,t)\in\mathcal E}|f_{i,t}-a_{i,t}|}{\sum_{(i,t)\in\mathcal E}a_{i,t}},
\label{eq:wape}
\end{equation}
for a forecast $f$. For a downstream step $A$ and an upstream step $B$, FVA is the relative error reduction
\begin{equation}
\mathrm{FVA}(A\,\|\,B)=100\cdot\frac{\mathrm{WMAPE}_B-\mathrm{WMAPE}_A}{\mathrm{WMAPE}_B},
\label{eq:fva}
\end{equation}
positive when $A$ improves on $B$. We report the RL policy against two upstream references, the statistical forecast and the stacked ML forecast it refines.

\begin{figure}[t]
\centering
\begin{tikzpicture}[
  box/.style={draw=black!70,rounded corners=2pt,align=center,font=\scriptsize,minimum height=8mm,text width=17mm,inner sep=2pt},
  arr/.style={-{Stealth[length=2mm]},thick,black!70},
  node distance=3.5mm,every node/.append style={}]
\node[box,fill=black!5] (f) {Features\\(Top drivers,\\Events)};
\node[box,fill=black!5,right=of f] (s) {AutoML Stacker\\(7 learners,\\top-3 mean)};
\node[box,fill=blue!8,right=of s] (a) {Bounds action set $\mathcal A$\\$\ell,P_{10..99},u,o$};
\node[box,fill=blue!15,right=of a] (r) {RL agent\\per-step reward\\$a^\star_{g,p}$};
\node[box,fill=orange!15,right=of r] (g) {Structural\\guard \&\\resolution};
\node[box,fill=black!12,below=3mm of g, text width=30mm] (p) {Published $=\hat y^{\mathrm{ML}}  +$agentic $+$planner};
\node[box,fill=black!5,below=3mm of s,text width=40mm] (h) {Context-matched replay of past periods\\(holidays, events, $S_p$, trend, price, promo)};
\draw[arr] (f)--(s); \draw[arr] (s)--(a); \draw[arr] (a)--(r); \draw[arr] (r)--(g); \draw[arr] (g)--(p);
\draw[arr] (h.east) -| (r.south);
\end{tikzpicture}
\caption{The ensemble learns the structure; the agent refines it. Each published value is one of the ensemble's own probabilistic values, chosen per period and checked by a structural guard.}
\label{fig:pipe}
\end{figure}
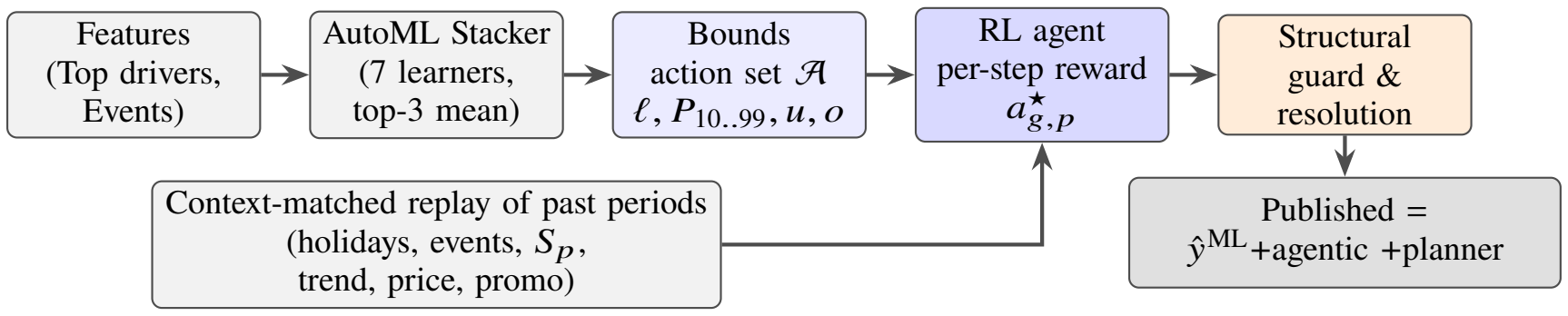

\section{AutoML Stacking and Distribution-Free Action Set}
\label{sec:actions}

\paragraph{Segmentation and cleaning.}
Series are segmented by ABC--XYZ classification (volume contribution and coefficient of variation) combined with a department descriptor; segments define donor pools for short histories and the granularity of decisions. For A and B clusters, one-off spikes are imputed to a robust centre and out-of-stock demand is reconstructed, since censoring otherwise teaches the model that demand fell when supply did.

\paragraph{Feature engineering.}
Ten feature families are generated and then screened: regressors and drivers (price, discount, spend, promotion and campaign flags, continuous drivers held apart from events), datetime and Fourier terms with a trend index, lags and offsets, encoded dimensions, per-holiday dummies and exogenous series. The lag grid is a dense block $\{1,\dots,6\}$ for short-run momentum plus custom lags $\{10,11,12,13,17,18,21,24\}$ that straddle the seasonal lag $m=12$, the half-cycle and the two-cycle lag. Rolling means over $\{1,2,3,12\}$ periods add a short and a full-cycle level, and lags respect the forecast gap to prevent leakage. Time-interaction dimensions let the seasonal shape differ across the product hierarchy. Mutual information $I(X_j;Y)$ is then estimated per candidate and the top $100$ features are kept, using a fractional threshold of $0.2$.

\paragraph{Learner pool.}
Seven learners span four inductive families: gradient-boosted trees (XGBoost, LightGBM)~\cite{ref_xgboost,ref_lightgbm}, a linear-booster XGBoost that extrapolates where trees cannot, a random forest that adds variance reduction with a different bias, and three neural models (MLP, LSTM, GRU), combined by stacked generalisation as in~\cite{ref_stacked}. Each is trained per segment and tuned with Tree-structured Parzen Estimators~\cite{ref_tpe,ref_hyperopt}.

The neural models capture price--promotion--calendar interactions, and the recurrent ones read lags as a gated sequence that can skip a stock-out gap or carry last year's peak, so their residuals are partly decorrelated from the trees'.

\paragraph{Stacking.}
Configurations are ranked on validation behaviour alone, so in-sample fit cannot win a place, and the best $k=3$ are combined as a convex stack,
\begin{equation}
\hat y_{i,t}=\sum_{j=1}^{k}\omega_i^{(j)}\hat y^{(j)}_{i,t},\qquad\omega_i^{(j)}\ge0,\quad\sum_j\omega_i^{(j)}=1.
\label{eq:stack}
\end{equation}
We use equal weights $\omega=1/k$, since fitted weights over a short validation window are high-variance~\cite{ref_m5acc}. The probabilistic values below are derived from the stack's residuals.

\paragraph{Chebyshev and Cantelli bounds.}
Let $r_{i,t}=a_{i,t}-\hat y_{i,t}$ over the validation window $\mathcal V_i$, with bias $b_i$ (mean residual) and dispersion $\sigma_i^2$ (residual variance about $b_i$). Demand residuals are skewed and heavy-tailed, so we use two inequalities that hold for any law with finite variance. Chebyshev bounds two-sided deviation, $\Pr(|Y-\mu|\ge\kappa\sigma)\le\kappa^{-2}$, so a tolerance $\alpha$ gives the symmetric multiplier $\kappa_\alpha=\alpha^{-1/2}$. Cantelli's one-sided refinement, $\Pr(Y-\mu\ge\kappa\sigma)\le(1+\kappa^2)^{-1}$, gives the tighter $\kappa^{C}_\alpha=\sqrt{(1-\alpha)/\alpha}$ for a single tail, and lets the two tails carry different tolerances $\alpha_\ell,\alpha_u$, which matches the asymmetric costs of overstock and stock-out. The lower and upper bounds are
\begin{equation}
\ell_{i,t}=\max\!\bigl(0,\hat y_{i,t}-\kappa_\ell\sigma_i\bigr),\qquad
u_{i,t}=\hat y_{i,t}+\kappa_u\sigma_i,
\label{eq:bounds}
\end{equation}
with Chebyshev multipliers by default and Cantelli multipliers when the tails are set separately.

\paragraph{Quantiles and offset.}
Percentile actions are running quantiles over the interval between the bounds,
\begin{equation}
P_q(i,t)=\ell_{i,t}+q\,\bigl(u_{i,t}-\ell_{i,t}\bigr),\qquad q\in\{0.1,\dots,0.9,0.99\},
\label{eq:ladder}
\end{equation}
which are monotone in $q$, an ordering the resolution step later relies on. A further action removes the measured bias rather than widening the interval, $o_{i,t}=\max(0,\hat y_{i,t}+b_i)$.

\paragraph{Action set.}
The thirteen sanctioned alternatives are
\begin{equation}
\mathcal A=\{\ell,\;P_{10},\dots,P_{90},P_{99},\;u,\;o\},
\label{eq:actions}
\end{equation}
and the base forecast is a fourteenth option. Write $v_a(i,t)$ for the value of action $a$ at row $(i,t)$. A policy is \emph{admissible} if it publishes only some $v_a(i,t)$, so every published figure is one the pipeline has already produced; this is relaxed in exactly one audited case (Sect.~\ref{sec:bound}). Each action has a plain reading in terms of the residual it corrects: $u$ and the upper rungs restore compressed peak amplitude or an under-weighted event, $\ell$ and the lower rungs deepen a compressed trough, and $o$ removes a persistent trend or level bias.

Concentration bounds are conservative ($\kappa^C_{0.1}=3$ against $1.64$ for a one-sided Gaussian $95\%$ bound), which is tolerable because the agent selects by measured reward and needs a well-ordered, sufficiently wide menu rather than calibrated quantiles.

\section{Attribution: Why Is the Forecast What It Is?}
\label{sec:factor}

To avoid double-counting effects the ensemble already carries, each forecast is decomposed before correction:
\begin{equation}
\hat y^{\mathrm{ML}}=B+T+S+E+D+M,
\label{eq:cake}
\end{equation}
with baseline $B$, trend $T$, seasonality $S$, events $E$, commercial drivers $D$ and a residual $M$, all non-negative and summing \emph{exactly} to the pre-correction output. The agent and the planner then appear as signed layers that are never clipped,
\begin{equation}
\mathrm{published}=\hat y^{\mathrm{ML}}+\underbrace{(\hat y^{\mathrm{RL}}-\hat y^{\mathrm{ML}})}_{\text{agentic}}+\underbrace{(\mathrm{pred}-\hat y^{\mathrm{RL}})}_{\text{planner}}.
\label{eq:layers}
\end{equation}
Layers come from a per-series surrogate: columns are routed to one bucket each (holidays always to events), a two-stage ridge fits events first, and credit is a counterfactual $\phi_j(t)=(x_j(t)-x^0_j(t))\beta_j$ against a family-specific reference (start of history for trend, cycle mean for seasonality, $0$ for a binary event, median for a continuous driver), rescaled to the model output. The decomposition enters the agent's state: it shows how much seasonality, trend and event lift the ensemble already carries, so an under-expressed peak can be told from one already learnt.

\section{Temporal Reinforcement Learning for Forecast Value Add}
\label{sec:rl}

\subsection{Decision Process}

We model refinement as a finite-action process $\langle\mathcal S,\mathcal A,R,\gamma\rangle$ that is \emph{temporal}: a decision is taken for each segment $g$ at each calendar position $p$ of the horizon, not once per series. The state $s_{g,p}$ holds the candidate values $\{v_a\}$, the calendar position, the structural seasonal index, event flags, growth diagnostics, the attribution layers and, centrally, the actuals, action values and commercial context of comparable past periods. An action is a choice $a\in\mathcal A$. In precise terms the selection is an offline contextual bandit evaluated by the direct method, without temporal credit assignment; we do not claim to solve a multi-step MDP.

\subsection{Per-Step Reward from Comparable Past Periods}
\label{sec:reward}

Absolute error alone favours a low, cautious forecast on spiky history, and matching the total alone tolerates months that are individually wrong, so the reward combines three components and is evaluated \emph{at each time step}, because the best action differs by period.

\paragraph{Context-matched replay.}
For horizon position $p$, reference periods are retrieved from the series' own history. Calendar alignment matches \emph{date minus $k$ years} with a frequency-aware tolerance (15 days monthly, 3 weekly, 1 daily), and moving events such as Black Friday, Easter or Lunar New Year are matched to their previous occurrence even in a different month. Each reference records the value $v_a(i,p_k)$ every action would have published against the realised $a_{i,p_k}$ and is weighted by how comparable its circumstances were,
\begin{equation}
\omega_k(i,p)\;\propto\;\rho^{k-1}\exp\!\Bigl(-\tfrac{1}{\tau}\bigl\|c_{i,p}-c_{i,p_k}\bigr\|^2\Bigr),
\label{eq:ctx}
\end{equation}
where $c_{i,p}$ stacks the holiday and event flags, the structural seasonal index, local trend growth, price index and promotion depth, and $\rho$ is a recency decay.

\paragraph{Reward components.}
For segment $g$ and position $p$, each component is higher-is-better:
\begin{align}
\mathrm{Acc}_{g,p}(a)&=1-\frac{\sum_{k}\sum_{i\in g}\omega_k\,|v_a(i,p_k)-a_{i,p_k}|}{\sum_{k}\sum_{i\in g}\omega_k\,a_{i,p_k}+\epsilon},\label{eq:acc}\\
\mathrm{Bias}_{g,p}(a)&=1-\frac{\bigl|\sum_{k}\sum_{i\in g}\omega_k\bigl(v_a(i,p_k)-a_{i,p_k}\bigr)\bigr|}{\sum_{k}\sum_{i\in g}\omega_k\,a_{i,p_k}+\epsilon},\label{eq:bias}
\end{align}
and $\mathrm{Adj}_{g,p}(a)$, which is \eqref{eq:acc} computed against de-spiked actuals, where observations outside an inter-quantile fence ($Q_{0.2}-1.5\,\mathrm{IQR}$, $Q_{0.8}+1.5\,\mathrm{IQR}$) are replaced by the series median. The step reward is their convex combination, and the step's winner is its maximiser:
\begin{equation}
R_{g,p}(a)=w_{\mathrm{acc}}\mathrm{Acc}_{g,p}+w_{\mathrm{bias}}\mathrm{Bias}_{g,p}+w_{\mathrm{adj}}\mathrm{Adj}_{g,p},\qquad
a^{\star}_{g,p}=\arg\max_{a\in\mathcal A}R_{g,p}(a),
\label{eq:reward}
\end{equation}
with weights $(0.34,0.33,0.33)$ in our deployments. The winner is then applied to the same position of the current horizon, taking the value $v_{a^\star}(i,p)$ of the current action set.

Because the reward is per step, the agent stays central where the ensemble already tracks the period and moves towards the correcting bound where the replay shows a residual, so a segment receives a \emph{profile of actions over the calendar} instead of one uniform shift, which is how seasonal amplitude, recurring events and one-off events are refined separately.

\subsection{Structural Guard on the Winning Action}
\label{sec:policyB}

A replayed winner is evidence from the past, so it is checked against an independent structural reconstruction. Series with under one cycle borrow seasonality from similar series (never making a row protected), and a series with no sales in the last three periods keeps the base forecast.

\paragraph{Structural seasonal profile.}
On a complete calendar grid, one seasonal shape is fitted across all cycles as a rank-one model $a_{y,p}\approx\ell_yS_p$ by alternating ratio estimates,
\begin{equation}
\ell_y=\frac{\sum_pa_{y,p}}{\sum_pS_p},\qquad
S_p=\frac{\sum_yw_y\,(a_{y,p}/\ell_y)}{\sum_yw_y},\qquad
w_y=\rho^{\,Y-y}\min\!\Bigl(1,\tfrac{c_y}{\bar c}\Bigr),
\label{eq:alt}
\end{equation}
where $c_y$ is the coverage of cycle $y$ and $Y$ the latest cycle; unlike the classical index, it stays unbiased when cycles differ in level or the live cycle is unfinished. Positions are classified from $S_p$, never from the base forecast: a \emph{peak} has $S_p\ge1.35$ (or is among the top two with $S_p>1.20$), a \emph{trough} has $S_p\le0.70$ (or is among the bottom two with $S_p<0.80$). A seasonality classifier (autocorrelation at the seasonal lag, peak-to-trough ratio, consistency of the last two cycles) sets how far the reconstruction leans on history: \emph{strong} seasonal series are rebuilt from a best-of-years shape times growth, \emph{weak} ones blend prior-cycle anchors with the base forecast, and \emph{non-seasonal} ones follow recent level and trend, with trivial edits skipped.

\paragraph{Structural target.}
The structural estimate is $F_{\mathrm{struct}}=L\cdot S_p\cdot E$, with a blended level $L$ and an event multiplier $E$ for an event recurring, moving in or moving away. A prior-cycle replay $F_{\mathrm{ly}}$ (calendar-aligned as above) uses a best-of-years anchor in high season,
\begin{equation}
A_{\mathrm{peak}}=\max\bigl(\mathrm{LY}_1(1+g_{\mathrm{peak}}),\;\mathrm{LY}_2(1+g_{\mathrm{up}})^2\,\theta_2\bigr),\qquad g_{\mathrm{up}}=\max(g,0),
\label{eq:anchor}
\end{equation}
with $\theta_2$ a trust factor on the older cycle, so one soft year cannot erase a spike the series has shown it can reach. The two are blended and the base forecast enters with a quality-weighted share:
\begin{align}
F_{\mathrm{hist}}&=w_sF_{\mathrm{struct}}+(1-w_s)F_{\mathrm{ly}},\label{eq:fhist}\\
\hat y^{\mathrm{unc}}&=(1-w_{\mathrm{ml}})F_{\mathrm{hist}}+w_{\mathrm{ml}}\hat y,\qquad w_{\mathrm{ml}}=w^{\mathrm{base}}_c(0.4+0.6\,\eta),\label{eq:unc}
\end{align}
with $\eta\in[0,1]$ the base model's in-sample quality and $w^{\mathrm{base}}_c$ set by seasonality class. On protected peaks the base share is at most $5\%$, and it is set to zero when $\hat y<0.5\,A_{\mathrm{peak}}$, that is, when the base horizon has not expressed the spike at all.

\paragraph{Protection and constraints.}
A protected peak is floored at $95\%$ of its anchor and a protected trough capped at $110\%$. Three constraints then apply: (i)~\emph{annual reconciliation}, scaling up uniformly any series implying more than $12\%$ de-growth against the calendar-matched prior year unless a decline is \emph{confirmed} (two consecutive negative cycles, confidence $\ge0.6$, growth $\le-15\%$), with unconfirmed negative growth floored at $-10\%$; (ii)~\emph{shape coherence}, smoothing the log-ratio to a probabilistic spine so its shape is borrowed without its level; and (iii)~a \emph{magnitude cap} of $2.0$ times the largest historical actual ($1.25$ for recent launches). The order is fixed, row targets, reconciliation, shape fit, cap, audits, resolution, guard-rail, and later steps override earlier ones. Beyond the first cycle, the corrected output becomes the synthetic prior year for the next, the one place where an action at $h$ alters the state at $h+m$.

\paragraph{Resolution.}
If $v_{a^\star}$ respects protection and the constraints it is published; otherwise the constrained target $\hat y^{\mathrm{unc}}$ is resolved to the nearest sanctioned action, once and without preference ordering,
\begin{equation}
a(i,t)=\arg\min_{a\in\mathcal A}\bigl|\hat y^{\mathrm{unc}}_{i,t}-v_a(i,t)\bigr|,
\label{eq:snap}
\end{equation}
with $\hat y^{\mathrm{unc}}$ and the distance exported on every row.

\paragraph{Bound guard-rail.}
\phantomsection\label{sec:bound}
One audited exception to admissibility is available and disabled by default. A row may exceed $u$ only if the series is in the top $80\%$ by horizon volume, the row is seasonal, the resolved value sits \emph{at} $u$ yet below the prior year, and the series is neither a trough row nor a confirmed decline. The target is $\min(1.05\,\mathrm{LY},\,2u,\,\mathrm{cap})$, and the action is labelled \texttt{upper\_bound\_x\%} with $x=100(\mathrm{target}/u-1)$ to flag the breach.

\subsection{Governance Properties}
\emph{Nothing is invented}: every figure is in $\mathcal A$ or the labelled exception. \emph{Nothing is absurd}: the cap bounds each row by its own history. \emph{Nothing is silent}: each row carries its action, unconstrained value, distance and flags. \emph{Nothing is unowned}: a named planner accepts or overrules, recorded as the planner layer of \eqref{eq:layers}.

\section{Worked Episode: One December Peak Row}
\label{sec:example}

We follow one decision through the loop of Sect.~\ref{sec:rl}, with illustrative parameters: a monthly, strong-seasonal SKU in the top $80\%$ by volume, guard-rail enabled, forecast for December 2026.

\paragraph{State.}
The structural index gives $S_{\mathrm{Dec}}=1.42$, so December is a protected peak. The matched references are December 2025 ($\mathrm{LY}_1=330$, soft after a stock-out) and December 2024 ($\mathrm{LY}_2=360$, the real spike). The ensemble forecast is $\hat y=160$: it has learnt that December is high, but the soft last year has pulled its level down. The largest historical actual is $400$, and growth parameters are $g_{\mathrm{peak}}=5\%$, $g_{\mathrm{up}}=4\%$, $\theta_2=0.90$.

\paragraph{Actions.}
The ensemble's probabilistic values for the row span $\ell=140$ to $u=280$; by \eqref{eq:ladder} the rungs are $P_{10}=154$, $P_{50}=210$, $P_{90}=266$ and $P_{99}=278.6$, and the bias-corrected value is $o=240$.

\paragraph{Reward.}
With recency decay $\rho=0.8$ and equal context distance, \eqref{eq:ctx} weights the two references $\omega=(0.556,0.444)$, a weighted actual of $343.3$. For brevity we let each reference year's backtested values coincide with the current ones. Every action then lies below both actuals, so all three components of \eqref{eq:reward} reduce to $v_a/343.3$ and the reward is monotone in the action value (Table~\ref{tab:episode}, top). The replay shows the December forecast has been persistently short; the winner is $a^\star=u$ with $R=0.816$.

\begin{table}[t]
\caption{The December episode. Top: per-step reward of selected actions from the context-matched replay. Bottom: the guard on the winning action and the published value.}\label{tab:episode}
\centering\scriptsize\setlength{\tabcolsep}{4pt}
\begin{tabular}{lccccccc}
\toprule
Action $a$ & $\ell$ & base $\hat y$ & $P_{50}$ & $o$ & $P_{90}$ & $P_{99}$ & $u$\\
Value $v_a$ & 140 & 160 & 210 & 240 & 266 & 278.6 & 280\\
Reward $R_{g,\mathrm{Dec}}(a)$ & 0.408 & 0.466 & 0.612 & 0.699 & 0.775 & 0.811 & \textbf{0.816}\\
\bottomrule
\end{tabular}

\smallskip
\begin{tabular}{rlll}
\toprule
\# & Step & Calculation & Result\\
\midrule
1 & Best-of-years anchor \eqref{eq:anchor} & $\max(330\times1.05,\ 360\times1.04^2\times0.90)$ & $A=350.4$\\
2 & Protection check on $a^\star$ & $v_u=280<0.95\times350.4=332.9$ & fails\\
3 & Peak target & $\hat y=160<0.5A=175.2\Rightarrow w_{\mathrm{ml}}=0$ & $350.4$\\
4 & Reconcile, shape, cap & $s=1$; peak keeps max; $350.4\le2\times400$ & $\hat y^{\mathrm{unc}}=350.4$\\
5 & Resolution \eqref{eq:snap} & nearest rung to $350.4$; distance $70.4$ & $u=280$\\
6 & Guard-rail test & top-80\%, peak, at $u$, YoY $=280/330-1=-15.2\%$ & fires\\
7 & Guard-rail target & $\min(346.5,\ 560,\ 800)$ & $346.5$\\
8 & Label & $x=(346.5/280-1)\times100$ & \texttt{upper\_bound\_23.8\%}\\
\bottomrule
\end{tabular}
\end{table}

\paragraph{Guard and publication.}
The winner moves in the right direction but not far enough: $280$ would ship a December $15.2\%$ below a stock-out year and violates peak protection. The guard recovers the spike from $\mathrm{LY}_2$ through \eqref{eq:anchor} and resolves it back to the ladder, again $u$. Because the top of the credible range lies below the prior year, the guard-rail fires and publishes $346.5$, $5\%$ above last year, labelled \texttt{upper\_bound\_23.8\%} so the breach is auditable (Table~\ref{tab:episode}, bottom). In the layers of \eqref{eq:layers} the published value is $160+186.5+0$: the ensemble's $160$, an agentic layer of $+186.5$ that restores the peak amplitude, and no planner override. The value then becomes the synthetic prior year for December 2027.

In short, the replay \emph{detects} a residual the ensemble left at a recurring calendar position and picks the direction; the guard \emph{sizes} it against the series' own structure; the action set and the label \emph{constrain} it to something a reviewer can trace.

\section{Experiments}
\label{sec:results}

\paragraph{Setup.}
We evaluate on five customer deployments: two retail (eggs; sweets), two consumer-packaged-goods (sports goods; cookware) and one direct-to-consumer (skincare). Difficulty ranges from sweets (WMAPE near $0.1$) to intermittent, event-driven sports goods (near $0.7$). Nine monthly forecasts, January to September 2026, are scored against realised demand for three methods: \textbf{Stat.}, the statistical forecast; \textbf{ML}, the stacked ensemble of Sect.~\ref{sec:actions}; and \textbf{RL}, the agent of Sect.~\ref{sec:rl} applied on top of ML with the same reward weights everywhere. The metric is WMAPE \eqref{eq:wape} for each customer-month. ``Window'' denotes the unweighted mean of the nine monthly WMAPEs, and the five-customer mean weights customers equally. FVA \eqref{eq:fva} is computed from these figures, and Win denotes the forecast with the lowest WMAPE in a customer-month.

\begin{table}[t]
\caption{WMAPE by customer and month (2026), rounded to two decimals; lower is better. The best forecast in each customer and column is in bold, and values tied after rounding are both in bold. Window is the mean of the nine months.}\label{tab:wmape}
\centering\scriptsize\setlength{\tabcolsep}{3.4pt}
\begin{tabular}{lcccccccccc}
\toprule
 & Jan & Feb & Mar & Apr & May & Jun & Jul & Aug & Sep & Window\\
\midrule
\multicolumn{11}{l}{\emph{Retail / Eggs}}\\
\quad RL & 0.27 & \textbf{0.24} & \textbf{0.26} & \textbf{0.23} & \textbf{0.22} & \textbf{0.29} & \textbf{0.25} & \textbf{0.28} & \textbf{0.21} & \textbf{0.25}\\
\quad ML & \textbf{0.26} & 0.27 & \textbf{0.26} & 0.25 & 0.25 & 0.33 & 0.28 & 0.31 & 0.25 & 0.27\\
\quad Stat. & 0.32 & 0.28 & 0.30 & 0.29 & 0.31 & 0.39 & 0.33 & 0.38 & 0.28 & 0.32\\
\multicolumn{11}{l}{\emph{Retail / Sweets}}\\
\quad RL & 0.11 & \textbf{0.09} & \textbf{0.11} & \textbf{0.09} & \textbf{0.09} & \textbf{0.12} & \textbf{0.10} & \textbf{0.11} & \textbf{0.09} & \textbf{0.10}\\
\quad ML & \textbf{0.10} & 0.11 & 0.14 & \textbf{0.09} & 0.12 & 0.16 & 0.12 & 0.14 & 0.11 & 0.12\\
\quad Stat. & 0.14 & 0.14 & 0.17 & 0.16 & 0.18 & 0.20 & 0.18 & 0.20 & 0.16 & 0.17\\
\multicolumn{11}{l}{\emph{CPG / Sports}}\\
\quad RL & 0.70 & \textbf{0.66} & \textbf{0.61} & 0.70 & \textbf{0.60} & \textbf{0.72} & \textbf{0.64} & \textbf{0.68} & \textbf{0.54} & \textbf{0.65}\\
\quad ML & \textbf{0.68} & 0.70 & 0.64 & \textbf{0.69} & 0.66 & 0.77 & 0.68 & 0.73 & 0.57 & 0.68\\
\quad Stat. & 0.78 & 0.71 & 0.71 & 0.79 & 0.72 & 0.86 & 0.75 & 0.82 & 0.64 & 0.75\\
\multicolumn{11}{l}{\emph{CPG / Cookware}}\\
\quad RL & 0.49 & \textbf{0.43} & \textbf{0.46} & 0.39 & \textbf{0.42} & \textbf{0.51} & \textbf{0.44} & \textbf{0.48} & \textbf{0.43} & \textbf{0.45}\\
\quad ML & \textbf{0.48} & 0.47 & 0.50 & \textbf{0.38} & 0.46 & 0.57 & 0.47 & 0.53 & 0.47 & 0.48\\
\quad Stat. & 0.56 & 0.51 & 0.56 & 0.48 & 0.54 & 0.64 & 0.55 & 0.60 & 0.52 & 0.55\\
\multicolumn{11}{l}{\emph{D2C / Skincare}}\\
\quad RL & \textbf{0.54} & 0.47 & \textbf{0.52} & 0.46 & \textbf{0.43} & \textbf{0.58} & \textbf{0.51} & \textbf{0.53} & \textbf{0.46} & \textbf{0.50}\\
\quad ML & 0.58 & \textbf{0.46} & 0.55 & \textbf{0.45} & 0.48 & 0.62 & 0.54 & 0.58 & 0.50 & 0.53\\
\quad Stat. & 0.63 & 0.54 & 0.62 & 0.55 & 0.56 & 0.70 & 0.61 & 0.66 & 0.57 & 0.60\\
\midrule
\multicolumn{11}{l}{\emph{Mean of the five customers}}\\
\quad RL & \textbf{0.42} & \textbf{0.38} & \textbf{0.39} & \textbf{0.37} & \textbf{0.35} & \textbf{0.44} & \textbf{0.39} & \textbf{0.42} & \textbf{0.35} & \textbf{0.39}\\
\quad ML & \textbf{0.42} & 0.40 & 0.42 & \textbf{0.37} & 0.39 & 0.49 & 0.42 & 0.46 & 0.38 & 0.42\\
\quad Stat. & 0.49 & 0.44 & 0.47 & 0.45 & 0.46 & 0.56 & 0.48 & 0.53 & 0.43 & 0.48\\
\bottomrule
\end{tabular}
\end{table}

\begin{table}[t]
\caption{Forecast Value Add of RL, in percent: how much lower the RL error was than the reference forecast. Negative values mean RL was worse. FVA is computed from unrounded WMAPE; the mean row uses the five-customer mean WMAPE of Table~\ref{tab:wmape}.}\label{tab:fva}
\centering\scriptsize\setlength{\tabcolsep}{2.6pt}
\begin{tabular}{lcccccccccc}
\toprule
 & Jan & Feb & Mar & Apr & May & Jun & Jul & Aug & Sep & Window\\
\midrule
\multicolumn{11}{l}{\emph{FVA of RL versus the stacked ML forecast}}\\
\quad Retail / Eggs     & $-3.8$ & 11.1 & $-2.0$ & 8.0 & 12.0 & 12.1 & 10.7 & 9.7 & 14.3 & 8.2\\
\quad Retail / Sweets   & $-5.9$ & 17.5 & 18.2 & $-4.7$ & 26.1 & 25.6 & 16.8 & 21.4 & 22.5 & 16.7\\
\quad CPG / Sports      & $-2.9$ & 5.7 & 4.7 & $-2.2$ & 9.1 & 6.5 & 5.9 & 6.8 & 5.3 & 4.3\\
\quad CPG / Cookware    & $-2.5$ & 7.5 & 8.0 & $-1.6$ & 8.7 & 10.5 & 6.4 & 9.4 & 7.5 & 6.3\\
\quad D2C / Skincare    & 6.9 & $-3.3$ & 5.5 & $-2.7$ & 10.4 & 6.5 & 5.6 & 8.6 & 8.0 & 5.3\\
\quad \textbf{Mean}     & $-0.4$ & 5.5 & 5.8 & $-0.9$ & 10.7 & 9.4 & 7.2 & 9.2 & 8.7 & \textbf{6.3}\\
\midrule
\multicolumn{11}{l}{\emph{FVA of RL versus the statistical forecast}}\\
\quad Retail / Eggs     & 15.6 & 14.3 & 13.3 & 20.7 & 29.0 & 25.6 & 24.2 & 26.3 & 25.0 & 21.9\\
\quad Retail / Sweets   & 21.7 & 34.7 & 34.9 & 43.8 & 51.4 & 40.8 & 44.7 & 45.0 & 44.9 & 40.8\\
\quad CPG / Sports      & 10.3 & 7.0 & 14.1 & 11.4 & 16.7 & 16.3 & 14.7 & 17.1 & 15.6 & 13.7\\
\quad CPG / Cookware    & 12.5 & 15.7 & 17.9 & 18.7 & 22.2 & 20.3 & 20.0 & 20.0 & 17.3 & 18.3\\
\quad D2C / Skincare    & 14.3 & 13.0 & 16.1 & 16.4 & 23.2 & 17.1 & 16.4 & 19.7 & 19.3 & 17.3\\
\quad \textbf{Mean}     & 13.2 & 13.3 & 16.9 & 17.6 & 23.9 & 20.5 & 19.8 & 21.8 & 20.3 & \textbf{18.7}\\
\bottomrule
\end{tabular}
\end{table}

\subsection{Results}

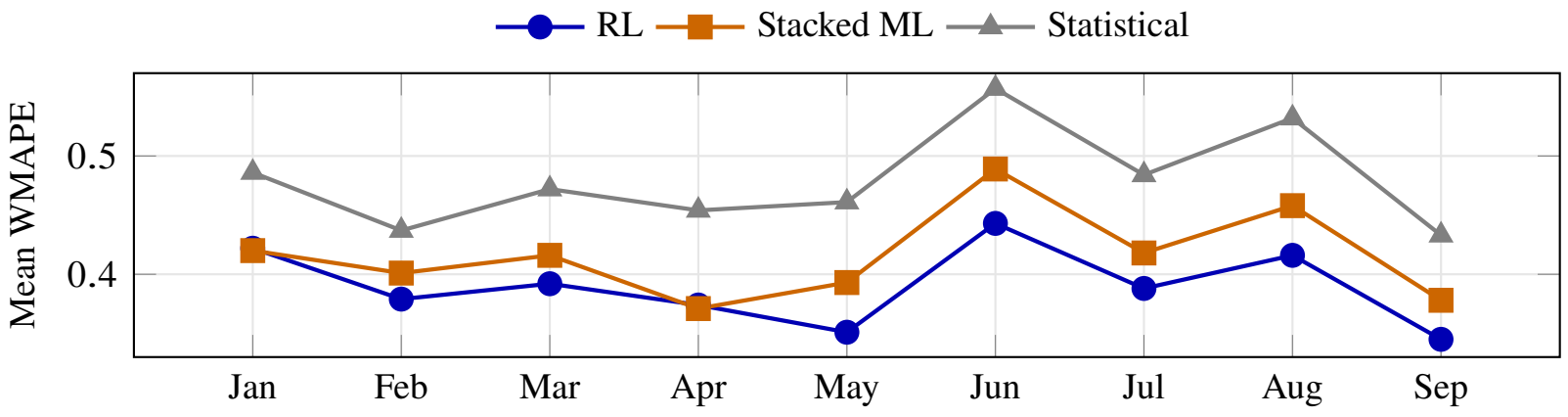
\begin{figure}[t]
\centering
\begin{tikzpicture}
\begin{axis}[
  width=0.92\textwidth, height=3.5cm,
  symbolic x coords={Jan,Feb,Mar,Apr,May,Jun,Jul,Aug,Sep},
  xtick=data, ymin=0.33, ymax=0.57,
  ylabel={Mean WMAPE}, ylabel style={font=\scriptsize},
  tick label style={font=\scriptsize},
  legend style={font=\scriptsize,at={(0.5,1.03)},anchor=south,legend columns=3,draw=none},
  grid=major, grid style={black!10}]
\addplot[mark=*,thick,blue!70!black] coordinates {(Jan,0.422)(Feb,0.379)(Mar,0.392)(Apr,0.374)(May,0.351)(Jun,0.443)(Jul,0.388)(Aug,0.416)(Sep,0.345)};
\addplot[mark=square*,thick,orange!80!black] coordinates {(Jan,0.420)(Feb,0.401)(Mar,0.416)(Apr,0.371)(May,0.393)(Jun,0.489)(Jul,0.418)(Aug,0.458)(Sep,0.378)};
\addplot[mark=triangle*,thick,black!50] coordinates {(Jan,0.486)(Feb,0.437)(Mar,0.472)(Apr,0.454)(May,0.461)(Jun,0.557)(Jul,0.484)(Aug,0.532)(Sep,0.433)};
\legend{RL,Stacked ML,Statistical}
\end{axis}
\end{tikzpicture}
\caption{Five-customer mean WMAPE by month. RL and ML coincide in January and April; the gap opens in the higher-error months of June and August.}
\label{fig:trend}
\end{figure}

Tables~\ref{tab:wmape} and~\ref{tab:fva} and Fig.~\ref{fig:trend} report the results.

\paragraph{Value added on top of the ensemble.}
Over the window, RL beats the stacked ML forecast for every customer, with FVA between $4.3\%$ (sports goods) and $16.7\%$ (sweets), and $6.3\%$ on the five-customer mean ($0.390$ against $0.416$). The ensemble is itself strong, $13.3\%$ better than the statistical forecast, so these gains are earned on top of a model that already learnt most of the structure. RL has the lowest WMAPE in $35$ of the $45$ customer-months ($78\%$) and the stacked ML forecast in the other $10$.

\paragraph{When the agent does not win.}
The ten ML wins are concentrated: four of the five customers in January, four in April, and one each in February (skincare) and March (eggs). In these months the agent's loss is small, between $1.6\%$ and $5.9\%$, and the five-customer mean is at parity ($-0.4\%$ and $-0.9\%$). Both follow a calendar break (year-end; an early Easter in 2026), where references are least comparable and the ensemble was already close. There a coarse menu can only cost resolution.

\paragraph{Where the value concentrates.}
From May to September RL wins all $25$ customer-months, and its largest gains over ML on the mean fall in May ($10.7\%$), June ($9.4\%$) and August ($9.2\%$). June and August are the highest-error months for all three forecasts (Fig.~\ref{fig:trend}), consistent with the residual being largest in peaks and event-heavy periods.

\paragraph{Consistency.} RL beats the statistical forecast in all $45$ customer-months (window FVA $13.7$--$40.8\%$, $18.7\%$ on the mean). Relative gains over ML are largest for the low-error sweets customer ($16.7\%$) and smallest for the high-error sports customer ($4.3\%$), but in absolute terms the order reverses: the reduction is $0.020$ WMAPE points for sweets and $0.029$ to $0.030$ for sports goods and cookware (Table~\ref{tab:wmape}, Window). Relative FVA therefore understates the value on hard categories.

\subsection{Limitations}

Five customers over nine months give breadth, not generality, and at most one instance of each peak. The agent does not win every month, and the window figures weight customers and months equally rather than by volume. Only accuracy is reported; a bias evaluation on the same customers is a natural next step. Per-step winners over-fit at series level, so we decide at segment level. Listing changes, confirmed orders and supply constraints remain the planner's responsibility.

\section{Conclusion}

We framed the refinement of a probabilistic ML forecast as reward-driven selection over the ensemble's own distribution-free values. The ensemble learns seasonality, trend and events; the agent scores each value at every time step against comparable past periods and moves the published value where the past shows a residual, refining seasonal amplitude, recurring and one-off events separately. Across five customer deployments the agent added value on top of a strong stacked ensemble for every customer, $6.3\%$ on average and up to $16.7\%$, was never worse than the statistical forecast, and was at parity with the ensemble in the months where it did not win. Future work includes a multi-step policy with temporal credit assignment and a learnt fallback to the central rung when replay evidence is weak.

\begin{credits}
\subsubsection{\ackname} The authors would like to thank Ankit Raj for his contributions towards RL architecture improvements; Ankit Solanki and Vasu Bhanderi for front development; Siddharth Shahi, Namrata Gupta and Sharad LM for helpful discussions and insights over the methodology and implementation.
\end{credits}

\end{document}